%% file: main.tex
\documentclass[twoside]{article}

\usepackage[accepted]{aistats2022}
\usepackage[round]{natbib}

\usepackage[utf8]{inputenc} 
\usepackage[T1]{fontenc}    

\usepackage[hyphens]{url}    
\usepackage{hyperref}       
\usepackage{booktabs}       
\usepackage{amsfonts}       
\usepackage{nicefrac}       
\usepackage{microtype}      
\usepackage{xcolor}         

\usepackage{graphicx}
\graphicspath{ {./images/} }
\usepackage{caption}
\usepackage{subcaption}

\usepackage{tikz}
\usetikzlibrary{bayesnet}

\usepackage{algorithm}
\usepackage{algpseudocode}

\begin{document}

%

%

\twocolumn[

\aistatstitle{ Marginalising over Stationary Kernels with Bayesian Quadrature}

\aistatsauthor{ Saad Hamid \And Sebastian Schulze \And Michael A. Osborne \And Stephen J. Roberts }

\aistatsaddress{ University of Oxford \And  University of Oxford \And University of Oxford \And University of Oxford } ]

\input{abstract}

\input{introduction}

\input{background}

\input{methods}

\input{relatedWork}

\input{experiments}

\subsubsection*{Acknowledgements}
The authors would like to thank anonymous reviewers for their detailed and constructive feedback.
S.H. acknowledges funding from EPSRC, and S.S. is supported by an I-CASE studentship funded by EPSRC and Dyson.

\bibliographystyle{plainnat}
\bibliography{references.bib}

\input{supplement_content}

\end{document}

%% file: abstract.tex
\begin{abstract}
Marginalising over families of Gaussian Process kernels produces flexible model classes with well-calibrated uncertainty estimates.
Existing approaches require likelihood evaluations of many kernels, rendering them prohibitively expensive for larger datasets.
We propose a Bayesian Quadrature scheme to make this marginalisation more efficient and thereby more practical.
Through use of the maximum mean discrepancies between distributions, we define a kernel over kernels that captures invariances between Spectral Mixture (SM) Kernels.
Kernel samples are selected by generalising an information-theoretic acquisition function for warped Bayesian Quadrature.
We show that our framework achieves more accurate predictions with better calibrated uncertainty than state-of-the-art baselines, especially when given limited (wall-clock) time budgets.
\end{abstract}

%% file: introduction.tex
\section{INTRODUCTION}
Gaussian Processes (GPs) \citep{rasmussen_gaussian_2006} are a rich class of models, which place probability distributions directly on classes of functions. Crucially, the success of these models is tied to the choice of their kernels. Analogously to architecture design in deep learning, kernels control the expressiveness and complexity of a GP model. If the correct kernel is chosen, GPs have shown the ability to make accurate predictions based on comparatively small training data sets. Beyond that, they natively provide predictive uncertainties with little additional computation. These confidence measurements can be vital for various down-stream tasks such as decision-making in the real world.

If, however, the chosen kernel is misspecified, little probability mass will be assigned to the neighbourhood of the true function, resulting in a poor fit. Many commonly used kernels, such as the Radial Basis Function and Mat\'ern kernels, have been shown to be universal kernels \citep{lugosi_universal_2006}, meaning they can approximate arbitrary functions given sufficient data. Despite this they, in practice, encode strong and frequently task-inappropriate inductive biases, significantly delaying learning progress.

To increase model flexibility and avoid such pathologies practitioners typically define parameterised \textit{kernel families}. The selection of a particular kernel is delayed until training and made data-dependent, e.g. via a maximum likelihood estimate (MLE).
A fully Bayesian approach gains further expressivity and robustness by marginalising across the chosen family instead of settling on a single kernel instance for test-time predictions. This is particularly valuable when working with large, complex data sets that exhibit complicated interactions between data points.

Our approach aims to create a flexible regression model capable of representing arbitrary stationary kernels and efficiently learning complex functions from large data sets. We draw on several previously separate strands of research to achieve this.

The analysis of kernels in the spectral domain has been shown to be an effective tool in learning expressive kernels \citep{wilson_gaussian_2013}. Due to \citet{bochner} a link between stationary kernel functions and frequency measures can be established. The former is a general class of kernels encompassing many popular families (e.g. RBF, Mat\'ern, periodic). The link to the latter allows the construction of posterior distributions in a principled and computationally efficient way with little user intervention and opens up clear avenues for their interpretation.

Rather than settling on a specific kernel via MLE, we conduct inference in the spectral-kernel framework by marginalising the entire kernel family against parameter likelihoods. The resulting integrals are computationally intractable and solutions can only be approximated. Previous approaches \citep{oliva_bayesian_2016, benton_function-space_2019, simpson_marginalised_2021} based on Monte Carlo variants effectively average across likelihood evaluations of kernels randomly sampled from the posterior. Each likelihood evaluation requires an inversion of the kernel Gram matrix over all data points, making such approaches increasingly expensive as we scale to larger data sets.  

To stay efficient even in large data, expensive likelihood settings, we adopt a model-based approach to integral approximation. Bayesian Quadrature (BQ) \citep{ohagan_bayeshermite_1991, rasmussen_bayesian_2003} methods model the function to be integrated directly, to incorporate and exploit prior knowledge of regularities (e.g. smoothness and non-negativity of likelihood surfaces \citep{osborne_active_2012}). This enables more careful sample acquisition, yielding performance superior to MCMC methods in wall-clock time for moderate-dimensional problems \citep{gunter_sampling_2014}.

Throughout this paper, we make the following contributions. 
Firstly, the construction of a custom \textit{hyper}-kernel based on the maximum mean discrepancy between distributions. This kernel is cheap to evaluate, whilst producing meaningful covariances between spectral densities, regardless of their parameterisation.
Secondly, a BQ framework based on this kernel to conduct computationally efficient GP inference on large data sets. 
Thirdly, the derivation of an information-theoretic acquisition function for the selection of new parameter evaluations.
Finally, we empirically demonstrate the viability of the resulting approach on several synthetic examples and real world data sets.
Our algorithm chooses informative likelihood evaluations and achieves high performance (log-likelihoods) on limited computational budgets measured in wall-clock time.

%% file: background.tex
\section{BACKGROUND}
\subsection{Gaussian Processes}
A Gaussian Process (GP) defines a probability distribution over the space of functions $f : \mathcal{X} \to \mathbb{R}$, such that for any finite subset $\mathcal{X}' \subset \mathcal{X}$, the vector $\{f(x)\}_{x \in \mathcal{X}'}$ is normally distributed.
We denote such a distribution $f \sim \mathcal{GP}\bigl( m, k \bigr)$, where the mean $m(x) = E[f(x)]$ and kernel functions $k(x_1,x_2)=E[(f(x_1)-m(x_1))(f(x_2)-m(x_2))]$ encode our prior beliefs about the function.
Analytic expressions for posterior mean and kernel of a process conditioned on a set of observations $D$ under a Normal likelihood are readily available.\footnote{A detailed discussion of the GP framework can be found in  \cite{rasmussen_gaussian_2006}.}

\subsubsection{Spectral Covariance Functions}
In this work the prior kernel functions $k(x_1,x_2)$ are assumed to be stationary. With slight abuse of notation, they take the form:
\begin{equation}
    k(x_1,x_2) = k(|x_1-x_2|) = k(\rho),
\end{equation}
such that two function values covary only depending on their separation $\rho$ and not their location.

Bochner's Theorem \citep{bochner} states that $k(\rho)$ is a positive-definite function on $\mathbb{R}$ (and valid kernel) if and only if its Fourier transform, $S(\omega)$, is a positive spectral density, i.e.:
\begin{equation}\label{bochner}
    k(\rho) = \int \text{e}^{2 \pi i \rho \omega} \text{d}S(\omega).
\end{equation}

As a consequence, any parameterisation of a set of spectral measures induces a corresponding parameterisation of stationary kernels.\footnote{For $k(\rho)$ to be real, we only consider symmetric $S(\omega)$.} 

\subsubsection{Kernels based on distance metrics}
Distance-based kernels which often take the form $k(x_1, x_2) = \lambda^2 \exp \bigl( - \frac{d(x_1, x_2)^q }{l^2} \bigr)$ ($\lambda$, and $l$ being hyper-parameters), are valid, i.e. positive definite, for $q=1$ if and only if the underlying metric $d(x_1, x_2)$ is conditionally negative definite \citep{jayasumana_kernel_2015,feragen_geodesic_2015}, meaning that it must give rise to matrices, $D$, such that $c^T D c < 0, \forall c : \sum_i c_i = 0$. For $q = 2$ the underlying metric must be Hilbertian, meaning that there must be an isometry between the metric space defining $d$ and a Hilbert space \citep{feragen_geodesic_2015}.

\subsection{Bayesian Quadrature}
Bayesian inference in machine learning frequently involves computation of intractable integrals of the form:
\begin{equation}
    Z = \int f(x)p(x) \text{d}x\quad,
\end{equation}
where $p(x)$ is a known prior density and $f(x)$ a (likelihood-)function. Bayesian quadrature is a model-based approach to approximately evaluating such integrals by modelling $f$ as a GP. Since GPs are closed under affine transforms the posterior over $Z$ is then Gaussian. For suitable kernel-prior combinations the quadrature weights can be evaluated analytically.

While the maintenance of the GP model requires some computational effort, it enables the generalisation of sample information across the integration domain. Consequently, samples can be chosen in a targeted fashion and BQ has been shown to be a competitive, more evaluation-efficient alternative to Monte Carlo methods, particularly in moderate-dimensional domains. 

Recent work \citep{osborne_active_2012,gunter_sampling_2014, chai_improving_2019}  further increases sample efficiency by encoding model-constraints such as the positivity of likelihood functions using warped GPs. This work builds upon WSABI \citep{gunter_sampling_2014} in particular, which places a GP prior over the square-root of $f$ and is described in more detail in Appendix \ref{WSABI}.

%% file: methods.tex
\section{OUR METHOD: MASKERADE}
In the following we present a flexible, data efficient Gaussian Process framework and show how to conduct Bayesian inference within it.

\subsection{The generative model}
Without loss of generality, we assume input dimensions of the training data to be rescaled to the interval $[0, 1]$ and outputs to be normalised to have zero mean and unit variance.

By this construction, a zero mean prior for our GP model is the most logical choice. As mentioned above, kernels are only assumed to be stationary. Following \citet{wilson_gaussian_2013} and \citet{oliva_bayesian_2016}, we parameterise spectral densities through Gaussian Mixture Models (GMMs) to induce a corresponding parameterisation of stationary kernels. In the limit of infinitely many components GMMs can approximate any spectral density to arbitrary precision.

The kernel, matching a spectral density parameterised by $\theta$, can be recovered via (\ref{bochner}). To create only real-valued kernels, we reflect all mixtures at the origin. We also note, that an additional output-scale $k(0)$ is required to cover the space of (unnormalised) measures and recover arbitrary stationary kernels. Since data has unit variance, however, we omit this in the following.

\begin{figure}[t]
    \centering
    \vspace{.4cm}
    \begin{tikzpicture}
    \node[obs] (DX) {$D_X$};
    \node[latent, right=of DX] (f) {$f$};
    \node[obs, right=of f] (DY) {$D_Y$};
    \node[latent, above=of f, yshift=-0.5cm] (k) {$k$};
    
    \node[latent, above=of k] (m) {$m$};
    \node[latent, left=of m] (w) {$w$};
    \node[latent, right=of m] (s) {$\sigma$};
    
    \node[latent, left=of w] (n) {$n$};
    \node[const, above=of n, yshift=-0.5cm] (N) {$N$};
    
    \node[const, above=of w, yshift=-0.5cm] (a) {$\alpha$};
    \node[const, above=of m, xshift=-0.5cm, yshift=-0.5cm] (mum) {$\mu$};
    \node[const, above=of m, xshift=0.5cm, yshift=-0.5cm] (sm) {$\Sigma$};
    \node[const, above=of s, xshift=-0.5cm, yshift=-0.5cm] (mus) {$\nu$};
    \node[const, above=of s, xshift=0.5cm, yshift=-0.5cm] (ss) {$\tau$};
    
    \edge {a} {w} ;
    \edge {mum,sm} {m} ;
    \edge {mus,ss} {s} ;
    
    \edge {N} {n};
    \edge {w,m,s} {k};
    \edge {k} {f};
    \edge {DX} {f};
    \edge {f} {DY};
    \plate {} {(w)(m)(s)} {$n$};
    \end{tikzpicture}
    \captionof{figure}{Bayesian Network of our model. The number of mixture components $n$ is drawn from a uniform prior; $\alpha$ parameterises a Dirichlet distribution over mixture weights $w$; $\mu$ and $\Sigma$ parameterise Gaussians over mixture means; $\nu$ and $\tau$ parameterise Log-Normal distributions over mixture scales. Not shown are hyperparameters of the hyper-kernel, $\lambda$ and $l$.\\}
    \label{bayes_net}
\end{figure}
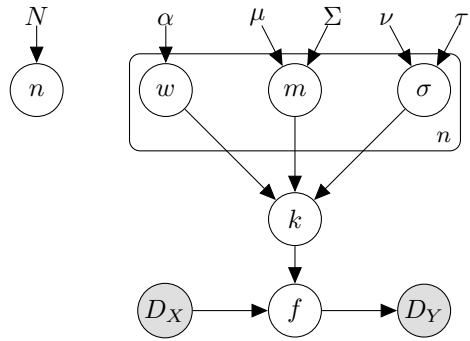

To perform Bayesian inference and marginalise across stationary kernels, we define a hyperprior $p(\theta)$ (see next section). A graphical representation of the resulting generative model is shown in Figure \ref{bayes_net}.

\subsubsection{ The Hyperprior $p(\theta)$}
\label{hyperprior}
The definition of our Bayesian model necessitates the choice of a suitable hyperprior. We see this is an advantage to the practitioner, as they are able to include existing knowledge to reduce the space of kernels needing  to be explored. In the absence of such information, we motivate a set of heuristics based on general properties of the training data. 

The prior over the number of mixture components $n$ is uniform up to some maximum $N$, limiting the size of the overall parameter space. The $n$ components are weighted against one another through a Dirichlet prior.

Mixture means are drawn from a normal distribution with zero mean and standard deviation $F_s / 5$ so that negligible mass lies above the Nyquist frequency $F_s$ of the data. Recall that the Nyquist frequency is the highest frequency directly identifiable from a data set.\footnote{For unevenly-spaced data we choose $F_s$ as the Nyquist frequency of a fictitious dataset with sampling frequency equal to the inverted average distance between datapoints.}

Mixture scales can be seen as approximate inverse lengthscales of the kernel. Accordingly their log-normal prior is set such that most of its mass lies between the inverse of the maximum distance between data points and the Nyquist frequency -- so that $\log(1 / |D_x|)$ (where $|D_x|$ is the size of the data window) is 5 standard deviations below the mean (of the normal distribution in log-space), and $\log(F_s)$ is 5 standard deviations above.

We stress that the prior outlined above is, in our view, the widest reasonable prior, and that performance is likely to be improved considerably by specifying narrower priors, if possible, especially for high dimensional datasets.

\subsection{Posterior Inference}
Since we do not know the full parameterisation of our GP model, we  marginalise out the unknown kernel parameters. Accordingly, the predictive distribution at some location $x_*$ under the above model, after observing data $D = \{x_D,y_D\}$, is given by:
\begin{equation} \label{infer}
\begin{split}
&p(y_* \mid x_*,D) = \int p(y_*\mid x_*, \theta, D)\, p(\theta \mid D) \text{d}\theta\\
&= \int p(y_*\mid x_*, \theta, D)\, \frac{p(D \mid \theta) p(\theta)}{p(D)} \text{d}\theta\\
&= \frac{ \int p(y_*\mid x_*, \theta, D)\, p(D \mid \theta)  p(\theta)
\text{d}\theta}{\int p(D \mid \theta)  p(\theta) \text{d}\theta},
\end{split}
\end{equation}

where $\theta = (n, \{w_i, m_i, \sigma_i\}_{i=1}^n)$ is the parameterisation of a spectral GMM with $n$ components, each described by a weight $w_i$, mean $m_i$ and scale $\sigma_i$. The predictive distributions $p(y_* \mid x_*, \theta,D)$ and likelihoods $p(D \mid \theta)$ for the  spectral kernel corresponding to a particular $\theta$ are given by the GP framework.

Unfortunately, both integrals in (\ref{infer}) are intractable and require approximation. We note that a major cost here is the evaluation of Gaussian Process likelihoods. These involve an inversion of the kernel Gram matrix of the entire data set whose computational cost increases quickly with data set size (typically cubically). To reduce the number of likelihood evaluations required, we propose to adopt a Bayesian Quadrature approach.

\subsection{Bayesian Quadrature}
We begin by constructing GP models of the functions $p(y_* \mid x_*,\theta,D)\,p(D \mid \theta)$ and $p(D \mid \theta)$ that are to be integrated against the prior. Since the locations $x^*$ of test points are unknown at training time, inference separates into two steps.

During training, a ``pseudo-dataset'' $\mathfrak{D}$ of decompositions and  matching likelihood evaluations (and model evidences respectively) is compiled through active sampling (see Section \ref{acquisition_function}) to shape our surrogate models and improve approximation quality. 
Hereby, we follow \citet{gunter_sampling_2014} and enforce positivity of the likelihood surrogate by placing a GP prior over the square-root of the likelihood function $L(\theta)=P(D \mid \theta)$:
\begin{equation}
\label{sqrt_transform}
    z = \sqrt{2(L(\theta) - \epsilon)} \sim \mathcal{GP}(0, \kappa),
\end{equation}
where $\kappa(\theta_1,\theta_2)$ is a kernel between spectral decompositions (see section \ref{sinkhorn_kernel}) and $\epsilon$ is a small non-negative constant.

Under this model $L(\theta)$ is not a GP, but can be approximated with one using a linear transformation of the surrogate. The posterior mean and covariance functions $\mathfrak{m}_\mathfrak{D}(\theta)$ and $\mathfrak{K}_\mathfrak{D}(\theta_1,\theta_2)$ then remain functions of the kernel $\kappa$ and the likelihood observations. To ensure a good fit to the actual likelihood function, hyperparameters are re-optimised after each evaluation. Since GPs are closed under continuous bounded linear transformations \citep{bogachev_gaussian_1961}, we obtain a Gaussian posterior over the model evidence -- the integral in the denominator of (\ref{infer}). The mean of this Gaussian takes the form $z^T Q z$, where $Q$ are the BQ weights -- a full derivation is given in Appendix \ref{WSABI}.

To make a prediction, we compute the integral in the numerator of \eqref{infer} in the same fashion. $p(y_* \mid x_*, \theta, D)$ is a GP posterior. Since we employ the same kernel $\kappa$ to measure similarity between spectral densities, and we reuse the same likelihood evaluations from the training stage, the quadrature weights are identical and only $z$ differs, allowing for efficient computation of the desired quantities. 

\subsubsection{The Hyper-Kernel}
\label{sinkhorn_kernel}
Noting that the densities under consideration are GMMs with a limited number of components, it becomes apparent that a suitable \textit{hyper}-kernel, $\kappa$, has to overcome two obstacles. Firstly, it needs to cope with varying numbers of parameters, or otherwise assign a predefined covariance between GMMs with differing numbers of components. Secondly, it should capture a number of invariances in the parameterisation of GMMs (e.g. reordering of components, subdivision or recombination of components etc.). A na\"ive approach for the construction of such a kernel would base distances between densities on the Euclidean distance between their parameters and fails to achieve either desiderata. 

Instead we construct a kernel based on the distance between the spectral densities represented by those parameters. Accordingly, we seek a conditionally-negative definite or Hilbertian metric between distributions. The maximum mean discrepancy meets our criteria, and is cheap to compute.

We obtain the following kernel:
\begingroup
\begin{equation}
\label{hyperkernel}
\kappa (\theta_1, \theta_2) = \lambda^2 \exp \left( - \frac{d(\theta_1, \theta_2)^q}{l^2} \right),
\end{equation}
\endgroup

where $\lambda > 0$ is an output-scale, $l > 0$ is a length-scale parameter, $q = \{1, 2\}$ and $d$ is a maximum mean discrepancy between the spectral densities (GMMs) parameterised by $\theta_1$ and $\theta_2$. The exact form of the MMD depends on the underlying kernel. Our experiments use the Energy Distance MMD \citep{feydy_geometric_2020}, given by
\begin{equation}
d(\theta_1, \theta_2) = w_1^T M_{12} w_2 - \frac{1}{2} w_1^T M_{11} w_1 - \frac{1}{2} w_2^T M_{22} w_2
\end{equation}
where $w_1$ and $w_2$ are vectors of component weights and $M_{12}$ is a matrix of euclidean distances between all pairs of component parameters $\{(m_i, \sigma_i)\}_{i=1}^{n_1}$ and $\{(m_i, \sigma_i)\}_{i=1}^{n_2}$, but our framework allows for alternatives (e.g. the Gaussian MMD, which can also be computed analytically for GMMs).
Note that, since the MMD is analytically computed (rather than approximated through sampling as is usual in the MMD literature), it is a valid distance between unnormalised densities.
The hyper-kernel therefore remains valid when kernels with arbitrary, non-unit output-scales are considered.

The only drawback is that integration of this kernel against the prior distribution is no longer possible analytically, so Monte Carlo integration is necessary. Arguably, however, the integration of the kernel against the prior lends itself better to such an approximation than that of the data likelihood. Intuitively, the likelihood function will have many sharp peaks for large datasets requiring samples to be drawn at very specific (unknown) locations for an accurate estimate. The kernel function by comparison is smoother, cheaper to evaluate and the resulting integral easier to approximate via random sampling from the prior. Our empirical results justify this approach, showing that we outperform competing methods despite the requirement for sampling.

\subsubsection{Information-theoretic point acquisition}
\label{acquisition_function}
The GP surrogate is trained based on a set of sampled spectral distributions, $\mathfrak{D}=\{(\theta_i, L_i)\}$, and the likelihood values of their corresponding kernels. Since adding a new point $\theta^\text{new}$ to this set requires a computationally expensive likelihood evaluation, it should be chosen to maximise its informativeness w.r.t. the quantity we care about -- the predictive distribution at test locations $x_*$.

As we do not expect the model to have access to the test locations at training time, we choose a different acquisition criterion for the new point. Instead, we aim to improve the approximation of the denominator of (\ref{infer}) -- and thereby the fit of the posterior distribution over spectral mixtures --  at training time. In \citet{gunter_sampling_2014} an acquisition function based on the uncertainty of the integrand is proposed. We extend the work of \citet{gessner20a} and \citet{fitbo} to propose an information-theoretic criterion instead.

We observe that the expected reduction in the entropy of our integral estimate after making an observation at sample location $\theta^s$ is given by
\begin{equation}
\label{acquisition_math}
\alpha(\theta_{*}) = H(L_{*} \mid \mathfrak{D},\theta_{*}) - E_{p(Z \mid \mathfrak{D},\theta_{*})}[H(L_{*} \mid \mathfrak{D},\theta_{*},Z)].
\end{equation}
Here $L_{*}$ is the predicted observation at location $\theta_{*}$ and $Z$ is the integral in the denominator of \eqref{infer}. Within the GP framework, the entropy of the predictive distribution $H(L_{*} \mid \mathfrak{D},\theta_{*})$ is the entropy of a Normal distribution and can be calculated in closed form. The second term turns out to be the expectation of a constant and the predictive distribution conditioned on $Z$ can once more be computed in closed form. \citet{gessner20a} discuss this acquisition function in the context of ordinary and multi-source Bayesian Quadrature. We note that it is available for warped Bayesian Quadrature. In particular, the acquisition function can be computed analytically when the linearisation approximation is used, provided that the kernel integrals are analytic. This is because the composition of the approximation and integration remains an affine transformation. Therefore, the integral of the approximate warped surrogate is jointly Gaussian distributed with the unwarped surrogate.

The acquisition function (\ref{acquisition_math}) is also valid for selecting batches of a fixed size. Importantly, our combination of hyper-kernel and acquisition function allows for the selection of (batches of) GMMs that are maximally informative about the overall evidence, $ \int p(D \mid \theta) p(\theta) \text{d}\theta$, rather than selecting GMMs for each parameter domain (corresponding to a fixed number of components) separately. This is because the kernel is able to define covariances between GMMs with different numbers of mixture components. Incorporating such information represents an improvement over \citet{chai_automated_2019} whose framework, in this instance, would be na\"ive as it assumes that the GPs over each domain are independent. 

\subsection{Computational Complexity}
During learning, each iteration requires optimising the hyperparameters of the surrogate GP and the acquisition function. The time complexity of hyperparameter optimisation is dominated by the cost of making likelihood evaluations, $\mathcal{O} (h^3)$ where $h$ is the number of Spectral Mixture kernels evaluated thus far. Acquisition function optimisation incurs an initial cost of $\mathcal{O} (h^3 + mh^2 + m^2)$, where $m$ is the number of Monte Carlo samples used to approximation the kernel integrals. Subsequent evaluations require $\mathcal{O} ( b h^2 )$ operations, where $b$ is the batch size. The memory complexity during hyperparameter optimisation is dominated by the cost of storing the hyper-kernel evaluated between observed SM kernels $\mathcal{O} (h^2)$. During the initialisation of the acquisition function, $\mathcal{O} (m^2)$ memory is additionally required to store the hyper-kernel evaluated between the MC samples.

Once the evaluation budget is exhausted the quadrature weights can be computed in $\mathcal{O} ( h^3 + mh^2 + m^2)$ time and $\mathcal{O}(h^2 + m^2)$ memory. Inference takes $\mathcal{O} (h |D|^3)$ operations, though this can be reduced significantly by disregarding terms that have low quadrature weight. The memory complexity of inference is $\mathcal{O} (h^2 |D|^2)$ for a na\"ive implementation, and can be reduced to $\mathcal{O} (h |D|^2)$ by looping appropriately (note that this leaves the asymptotic time complexity unchanged).

%% file: relatedWork.tex
\section{RELATED WORK}
\subsection{Kernel learning}
Early work in kernel learning focussed on searching over compositions (sums and products) of a set of basic covariance functions \citep{duvenaud_structure_2013}. 
More recent work has proposed learning kernels using their spectral representations \citep{wilson_gaussian_2013,samo_advances_2017,lazaro-gredilla_sparse_2010,gal_improving_2015,samo_generalized_2015,oliva_bayesian_2016,remes_non-stationary_2017,jang_scalable_2017,ambrogioni_integral_2018,tobar_bayesian_2018,teng_scalable_2019,benton_function-space_2019}.
Beyond the ability to capture more complex structures, this approach also lends itself to approximate inference, reducing learning time complexity \citep{gal_improving_2015,rahimi_random_2008,hensman_variational_2018}. A model similar to ours uses a Dirichlet Process prior over the number of components in a Gaussian Mixture Model (GMM) residing in the spectral domain \citep{oliva_bayesian_2016}.
\citet{simpson_marginalised_2021} use Nested Sampling to marginalise over the parameters of a Spectral Mixture kernel with a fixed number of mixture components.

Inference in spectral models reduces to the marginalisation of the likelihood function against the posterior over kernels (or, equivalently, their spectral decompositions). The resulting integrals are intractable and have to be approximated -- usually through the use of an appropriate MCMC scheme, such as Gibbs sampling, RJ-MCMC \citep{green_reversible_1995} or elliptical slice sampling \citep{murray_elliptical_2010}. Scaling MCMC marginalisation over the spectral kernel parameters up to large datasets is problematic. The computational cost of likelihood evaluations scales poorly in the size of the training data set and the collection of sufficiently many samples for an accurate Monte-Carlo estimate becomes prohibitively expensive. As above, BQ may be a more appropriate choice in these settings.

The recently proposed baselines we will compare against in our experiments include:
\begin{description}
\item [\textit{Variational Sparse Spectrum Gaussian Process}] \citet{gal_improving_2015} which performs Variational Inference over a sparse spectrum approximation.
\item [\textit{Bayesian Nonparametric Kernel Learning}] \citet{oliva_bayesian_2016} which learns the posterior distribution over Spectral Mixture kernels via a parameterisation of spectral densities based on GMMs and a Gibbs sampling scheme.
\item [\textit{Functional Kernel Learning}] \citet{benton_function-space_2019}  which places a GP prior over the kernel's spectral density and infers a posterior over kernels using an Elliptical Slice Sampling scheme.
\end{description}

Various approximations to reduce the cost of likelihood evaluations have been proposed in the literature, such as PCG \citep{cutajar2016} and  BBMM \citep{gardner_gpytorch_2018} -- modified conjugate gradient methods -- and Random Fourier Features \citep{rahimi_random_2008}. This is orthogonal to our investigation and our framework is optionally capable of leveraging both of these for scalability.

\subsection{Bayesian Quadrature}
Recent work in BQ \citep{osborne_active_2012, gunter_sampling_2014, chai_improving_2019} has proposed the use of warped GPs \citep{snelson_warped_2004} to incorporate a priori known model-constraints into the surrogate. 
This work builds upon WSABI \citep{gunter_sampling_2014} in particular, which places a GP prior over the square-root of the integrand.
The acquisition function developed in Section~\ref{acquisition_function} also applies to this setting and is an extension of the information-theoretic scheme discussed in \citet{gessner20a}. \citet{osborne_bayesian_2012} also proposed a BQ framework to infer ratios of integrals, which is relevant to the computation of marginalised posteriors.
However, the combination with WSABI and our framework introduces further intractable integral terms, which must be approximated with Monte Carlo sampling and would prohibitively raise the cost of inference.
Additionally, the work of \citet{xi_bayesian_2018} and \citet{chai_automated_2019} is related as it concerns the use of BQ for computing correlated integrals, and for automatic model selection.

\subsection{Gaussian Processes on Spaces of Measures}
The BQ integrand model requires the specification of a positive-definite kernel across the integration domain -- the space of spectral densities. While numerous metrics and divergences have been proposed to measure differences between probability distributions, many do not give rise to valid kernels. 

Recent work has used Wasserstein distances \citep{bachoc_gaussian_2018} to define Gaussian Processes on a space of measures.
However, Wasserstein distances are only Hilbertian in 1D \citep{peyre_computational_2019} so extensions to multidimensional distributions rely on Hilbert space embeddings of optimal transport maps to a reference distribution \citep{bachoc_gaussian_2019}.
Furthermore, evaluating Wasserstein distances can be expensive as it involves finding the solution to an optimal transport problem.
The Independence kernel based on the Sinkhorn distance \citep{cuturi_sinkhorn_2013} has been shown to be a valid positive definite kernel, but the distance between identical distributions may not be zero.
An elegant alternative that satisfies our desiderata are maximum mean discrepancies (MMDs), which are defined in terms of kernel mean embeddings.
MMDs are Hilbertian, so are a valid metric for kernels of the form \eqref{hyperkernel} \citep{muandet_kernel_2017}.

%% file: experiments.tex
\section{RESULTS}
\label{experiments}
In the following we empirically assess the performance of our model on a variety of tasks. 

\subsection{Experiment setup}
The experiments were conducted on Nvidia Titan V\footnote{Comparisons to BaNK and VSSGP, and ablation studies.} and Nvidia GeForce GTX 1080\footnote{Comparison to FKL.} GPUs. We report the Root Mean Squared Errors (RMSE) of the mean function of the predictive posterior as well as the Log-Likelihoods (LL) of the predictive posterior on held out test data after training each model for a fixed training time budget. Since the cost of likelihood evaluations depends on the dataset size, so does the training budget. The numbers given indicate either mean performance or mean performance and standard deviation. 

We compare our approach of MArginalising Spectral KERnels As DEnsities (MASKERADE)\footnote{https://github.com/saadhamidml/maskerade} to models previously proposed in the literature including VSSGP \citep{gal_improving_2015}, BaNK \citep{oliva_bayesian_2016}, and FKL \citep{benton_function-space_2019}. Results for the first two algorithms are directly taken from the papers. FKL hyperparameters were set to the defaults for the \texttt{spectralgp}\footnote{https://github.com/wjmaddox/spectralgp} Python package, with the $\omega_{max}$ (maximum frequency) argument modified to match the Nyquist frequency of the dataset. 

We limit the parameter space MASKERADE considers to GMMs with up to 5 components (N=5). The concentration of the Dirichlet prior over weights is set to $\alpha = 1$. The remaining priors are chosen as described in section \ref{hyperprior}. We initialise the BQ surrogate GPs with likelihood evaluations of parameters randomly chosen from the prior and thereafter acquire additional evaluations using the acquisition function. LBFGS is used to optimise both the acquisition function as well as the parameters of the hyper-kernel.

\subsection{Medium scale data sets}
We begin by fitting MASKERADE to the Solar dataset \citep{lean_solar_2004} to compare against VSSGP. We replicate the setup from \citet{gal_improving_2015}, withholding 5 sets of length 20 as the test set.
\begingroup
\begin{table}[h]
\caption{Test set RMSE for the Solar Irradiance Data Set. Results for VSSGP are taken from \citet{gal_improving_2015}.}
\vskip 0.15in
\begin{center}
\begin{small}
\begin{sc}
\begin{tabular}{lcccc}
\toprule
 & VSSGP & MASKERADE \\
\midrule
RMSE & 0.41 & \textbf{0.17} \\
\bottomrule
\end{tabular}
\end{sc}
\end{small}
\end{center}
\vskip -0.1in
\end{table}
\endgroup

We further examine the algorithms' performance on four datasets from the UCI Machine Learning Repository \citep{Dua:2019}: Yacht Hydrodynamics (308 instances, 6 input dimensions) (YH) \citep{yacht}, Auto MPG (MPG) (398 instances, 8 input dimensions) \citep{autompg}, Concrete Compressive Strength (CCS) (1030 instances, 8 input dimensions) \citep{concrete}, and Airfoil Self-Noise (ASN) (1503 instances, 5 input dimensions) \citep{airfoil}.

To compare against FKL we follow \citet{benton_function-space_2019} and average over 10 runs. We use 10-fold cross validation so that the tests sets remain independent, and the statistical significance of the results can be assessed using paired Student-t tests. These results are presented in Table ~\ref{tab:fkl_ll}.

\begin{table}[h]
\caption{Posterior log likelihood of and RMSE on the test set for UCI MLR datasets. Results for FKL taken from \citet{benton_function-space_2019}.}
\vskip 0.15in
\begin{center}
\begin{small}
\begin{sc}
\begin{tabular}{lccc}
\toprule
& \multicolumn{2}{c}{LL} & P-value \\
\cmidrule(lr){2-3} \cmidrule(lr){4-4}
 & FKL & MASKERADE \\
\midrule
YH & -32.1 $\pm$ 8.45 & \textbf{47.2 $\pm$ 5.76} & 3 $\times$ 10\textsuperscript{-9} \\
MPG & -104 $\pm$ 5.64 & \textbf{-57.3 $\pm$ 3.95} & 4 $\times$ 10\textsuperscript{-9} \\
CCS & -301 $\pm$ 34.1 & \textbf{-123 $\pm$ 6.49} & 1 $\times$ 10\textsuperscript{-7} \\
ASN & -864 $\pm$ 36.0 & \textbf{-152 $\pm$ 13.3} & 2 $\times$ 10\textsuperscript{-13} \\
\bottomrule\\
& \multicolumn{2}{c}{RMSE} & P-value \\
\cmidrule(lr){2-3} \cmidrule(lr){4-4}
 & FKL & MASKERADE \\
\midrule
YH & \textbf{0.63 $\pm$ 0.36} & 2.01 $\pm$ 0.58 & 1 $\times$ 10\textsuperscript{-6} \\
MPG & \textbf{3.09 $\pm$ 0.36} & 8.14 $\pm$ 0.66 & 7 $\times$ 10\textsuperscript{-9} \\
CCS & \textbf{5.81 $\pm$ 2.65} & 13.1 $\pm$ 1.31 & 3 $\times$ 10\textsuperscript{-5} \\
ASN & 83.9 $\pm$ 12.3 & \textbf{4.59 $\pm$ 0.62} & 1 $\times$ 10\textsuperscript{-8} \\
\bottomrule
\end{tabular}

\end{sc}
\end{small}
\end{center}
\label{tab:fkl_ll}
\vskip -0.1in
\end{table}

Similarly, to compare against BaNK we follow \citet{oliva_bayesian_2016} and perform 3 repeats of 5-fold cross-validation on the Concrete Compressive Strength and Airfoil Self-Noise datasets, for which we present results in Table ~\ref{tab:bank}.

\begin{table}[h]
\caption{Test set MSE on two UCI data sets. Results for BaNK taken from \citet{oliva_bayesian_2016}.}
\vskip 0.15in
\begin{center}
\begin{small}
\begin{sc}
\begin{tabular}{lcc}
\toprule
 & BaNK & MASKERADE \\
\midrule
CCS &  \textbf{0.1195 $\pm$ 0.0108} & 0.6889 $\pm$ 0.0168 \\
ASN & 0.3359 $\pm$ 0.0354 & \textbf{0.2578 $\pm$ 0.0866} \\
\bottomrule
\end{tabular}
\end{sc}
\end{small}
\end{center}
\label{tab:bank}
\vskip -0.1in
\end{table}

\subsection{Large scale data sets}
Finally we compare MASKERADE against FKL on two large datasets, with a limited time budget of 20 minutes for learning.
We again average over 10 runs and report results in Table ~\ref{tab:time_ll}. MASKERADE once more achieves superior performance to FKL on the same time budget.
The datasets are the Sterling Broad Based Exchange Rate (SER) \citep{sterling} (data from 1990-01-02--2020-10-08, totalling 7781 data points) and the UCI 3D Road Network (RN) \citep{road} (10,000 data point subset of ~250,000 data points. We choose the first 10,000 instances with a unique OSM-ID) for which the task is to predict altitude from latitude and longitude.

\begin{table}[h]
\caption{Posterior log likelihood of and RMSE on the test set for FKL and MASKERADE fit to two large datasets. Both methods were given a time budget of 20 minutes for training.}
\vskip 0.15in
\begin{center}
\begin{small}
\begin{sc}
\begin{tabular}{lccc}
\toprule
& \multicolumn{2}{c}{LL} & \multicolumn{1}{c}{P-value} \\
\cmidrule(lr){2-3} \cmidrule(lr){4-4}
 & FKL & MASKERADE\\
\midrule
SER & -1054 $\pm$ 30.70 & \textbf{335.4 $\pm$ 53.58} & 8 $\times$ 10\textsuperscript{-8} \\
RN & -4355 $\pm$ 219.8 & \textbf{-1071 $\pm$ 21.55} & 7 $\times$ 10\textsuperscript{-12} \\
\bottomrule\\
& \multicolumn{2}{c}{RMSE} \\
\cmidrule(lr){2-3}
DS & FKL & MASKERADE\\
\midrule
SER & 1.042 $\pm$ 0.028 & \textbf{0.523 $\pm$ 0.042} & 2 $\times$ 10\textsuperscript{-10} \\
RN & \textbf{10.17 $\pm$ 0.62} & 12.80 $\pm$ 0.427 & 3 $\times$ 10\textsuperscript{-6} \\
\bottomrule\\
\end{tabular}
\end{sc}
\end{small}
\end{center}
\label{tab:time_ll}
\vskip -0.1in
\end{table}

\subsection{Ablation Study}
We conduct an ablation study to verify the effectiveness of two components of our model in Table ~\ref{table:ablation}.
Firstly, we verify that marginalising across the kernel family does indeed lead to better predictive performance. As a baseline, we compare against a GP using Spectral Mixture kernel (SM) whose hyperparameters were optimised via gradient descent to maximise the likelihood of the training data.
Secondly, we compare our proposed acquisition function against uncertainty sampling \citep{gunter_sampling_2014} and random point acquisition. We set an evaluation budget of 1000 and evaluate performance on the Airline Passenger (AP) \citep{airpass} and Mauna Loa Atmospheric CO\textsubscript{2} Concentration (ML) \citep{maunaloa} datasets. We average over 10 runs with a 90/10 train/test split.
In both cases, our chosen approach compares favourably to alternative designs.

\begin{table}
\caption{Test set RMSE for the Airline Passenger and Mauna Loa datasets. MASKERADE-I indicates the use of our proposed information theoretic acquisition function, MASKERADE-U indicates uncertainty sampling \cite{gunter_sampling_2014}, and MASKERADE-R indicates random sampling under the prior.}
\vskip 0.15in
\begin{center}
\begin{small}
\begin{sc}
\begin{tabular}{lcc}
\toprule
 & SM & MASKERADE-I\\
\midrule
AP &  0.283 $\pm$ 0.000 & \textbf{0.162 $\pm$ 0.020}\\
ML & 0.802 $\pm$ 0.685 & \textbf{0.026 $\pm$ 0.002}\\
\bottomrule\\
\toprule
 & MASKERADE-U & MASKERADE-R \\
\midrule
& \textbf{0.164 $\pm$ 0.027} & \textbf{0.163 $\pm$ 0.022} \\
& 0.029 $\pm$ 0.001 & 0.030 $\pm$ 0.002 \\
\bottomrule
\end{tabular}
\end{sc}
\end{small}
\end{center}
\vskip -0.1in
\label{table:ablation}
\end{table}

\section{DISCUSSION}
\label{discussion}
We have introduced a novel framework for kernel learning for Gaussian Processes that seeks to marginalise over Spectral Mixture Kernels using Bayesian Quadrature.
Specifically we use a maximum mean discrepancy as a metric underlying an exponential kernel to define a Gaussian Process on a space of GMMs, and show how this can be efficiently computed.
This elegantly incorporates invariances between Spectral Mixture Kernels into our model.
Additionally we show that an information-theoretic acquisition function is applicable for warped Bayesian Quadrature, and how to use it within our framework.
We empirically evaluate our method on several datasets and find that it is competitive with state-of-the-art baselines.

The key limitation of our method is due to the curse of dimensionality.
The most scalable instantiation of our framework handles multi-dimensionality by using GMMs in which each Gaussian has a diagonal covariance structure.
This means that the number of parameters for an SM kernel with $M$ mixtures is $M(1 + 2D)$, which grows faster than the number of data dimensions, $D$.
This limits the effectiveness of any Bayesian Quadrature regime.
Note that Monte Carlo methods also suffer in high dimensional spaces, especially if likelihood evaluations are expensive.
(Our method relies on MC integration of kernel integrals, but these are considerably easier to compute than the marginalisation integrals because our method only requires samples from the priors over hyperparameters rather than the posteriors, and the kernel is much cheaper to evaluate than the likelihood.)
Our approach is most beneficial for inference based on large, low dimensional datasets.

The societal impacts of this work will depend on the problems that practitioners apply it to, as it is a general-purpose framework.

%% file: supplement_content.tex
\clearpage
\appendix

\thispagestyle{empty}

\onecolumn \makesupplementtitle

\section{SUMMARY OF WSABI}\label{WSABI}
For completeness we include a summary of the Warped Sequential Active Bayesian Integration (WSABI) algorithm proposed by \citet{gunter_sampling_2014}, upon which our work builds heavily.

Recall that we are interested in computing integrals of the form $\int f(\theta) \mathrm{d}\pi(\theta)$, where $f(\theta)$ is non-negative.
We enforce this constraint by modelling $g(\theta) = \sqrt{2 \bigl( f(\theta) - \epsilon \bigr) } \sim \mathcal{GP} (0, \kappa)$.
This induces a non-central Chi-squared distribution over $f(\theta)$.

Now, denote by $\Theta$ the set of observation locations, the matching transformed observations by $z = \sqrt{2 \bigl( f(\Theta) - \epsilon \bigr) }$, the covariance between all pairs of observations as $\kappa_{\Theta\Theta}$, the covariance between an arbitrary point $\theta$ and the set of observations as $\kappa_{\Theta\Theta}$ and the kernel between two arbitrary points as $\kappa_{\theta_1\theta_2}$.
Additionally, let $\mathfrak{D} = (\Theta, z)$.

Further, recall that the posterior over $g(\theta)$ conditioned on $z$ is given by $\mathcal{GP} (\mu_{\mathfrak{D}}, \Sigma_{\mathfrak{D}})$ with:
\begin{align*}
\mu_{\mathfrak{D}} &= \kappa_{\theta\Theta} \kappa_{\Theta\Theta}^{-1} z\\
\Sigma_{\mathfrak{D}} &= \kappa_{\theta_1\theta_2} - \kappa_{\theta_1\Theta} \kappa_{\Theta\Theta}^{-1} \kappa_{\theta_2\Theta}^T
\end{align*}

By taking a Taylor expansion around the posterior mean of $g(\theta)$, $\mu_{\mathfrak{D}}$, we can approximate
\begin{align*}
	f ( g ) &\approx f(\mu_{\mathfrak{D}}) + (g - \mu_{\mathfrak{D}}) \frac{\mathrm{d}f}{\mathrm{d}g} \biggr|_{g = \mu_{\mathfrak{D}}} \\
	&= \epsilon - \frac{1}{2} \mu_{\mathfrak{D}}^2 + \mu_{\mathfrak{D}} g.
\end{align*}
As this is a linear transform of $g$, we can approximate the distribution over $f$ as $\mathcal{GP} (\mathfrak{m}_{\mathfrak{D}}, \mathfrak{K}_{\mathfrak{D}})$ with moments
\begin{align*}
	\mathfrak{m}_{\mathfrak{D}} (\theta) &= \epsilon + \frac{1}{2} \mu_{\mathfrak{D}}(\theta)^2 , \\
	\mathfrak{K}_{\mathfrak{D}} (\theta_1, \theta_2) &= \mu_{\mathfrak{D}}(\theta_1) \Sigma_{\mathfrak{D}}(\theta_1, \theta_2) \mu_{\mathfrak{D}}(\theta_2).
\end{align*}

Then the first two moments of the integral of interest are given by
\begin{align*}
	\mathbb{E} \biggl[ \int f(\theta) \mathrm{d}\pi(\theta) \biggr] &=  \int \mathbb{E}[f(\theta)] \mathrm{d}\pi(\theta) = \int \biggl( \epsilon + \frac{\mu_{\mathfrak{D}}(\theta)^2}{2} \biggr) \mathrm{d}\pi(\theta) \\
	&= \epsilon + \frac{1}{2} \biggl( z^T \kappa_{\Theta\Theta}^{-1} \int \kappa_{\theta\Theta}^T \kappa_{\theta\Theta} \mathrm{d}\pi(\theta) \kappa_{\Theta\Theta}^{-1} z \biggr) \\
	&= \epsilon + \frac{1}{2} z^T Q z.
\end{align*}
and
\begin{align*}
    \mathrm{Cov} \biggl[ \int f(\theta) \mathrm{d}\pi(\theta) \biggr] &= \iint \mu_{\mathfrak{D}}(\theta) \Sigma_{\mathfrak{D}}(\theta, \theta') \mu_{\mathfrak{D}}(\theta') \mathrm{d}\pi(\theta) \mathrm{d}\pi(\theta') \\
    &= z^T \kappa_{\Theta\Theta}^{-1} \biggl( \iint \kappa_{\theta\theta'} \kappa_{\theta\Theta}^T \kappa_{\theta'\Theta} \mathrm{d}\pi(\theta) \mathrm{d}\pi(\theta') - \int \kappa_{\theta\Theta}^T \kappa_{\theta\Theta} \mathrm{d}\pi(\theta) \kappa_{\Theta\Theta}^{-1} \int \kappa_{\theta'\Theta}^T \kappa_{\theta'\Theta} \mathrm{d}\pi(\theta') \biggr) \kappa_{\Theta\Theta}^{-1} z\\
\end{align*}

Where the quadrature weights, $ Q = \kappa_{\Theta\Theta}^{-1} \int \kappa_{\theta\Theta}^T \kappa_{\theta\Theta} \mathrm{d}\pi(\theta) \kappa_{\Theta\Theta}^{-1}$, only depend on the particular kernel choice and locations of already observed function values.

For the numerator of Equation (4) in the main text we simply substitute $z_* = \sqrt{2 \bigl( p(y_* \mid x_*, D, \theta) p(D \mid \theta) - \epsilon \bigr)}$ for $z = \sqrt{2 \bigl( p(D \mid \theta) - \epsilon \bigr)}$. The predictive posterior is then a weighted sum of products of Gaussians.

\section{A REMARK ON POSTERIOR INFERENCE \eqref{infer}}
The predictive posterior given in \eqref{infer} is typically viewed through the heirarchical Bayesian framework:
\begin{equation}
\label{}
\begin{split}
p(y_* \mid x_*,D) &= ... = \frac{ \int p(y_*\mid x_*, \Theta,{} D)\, p(D \mid \Theta)  p(\Theta)
\text{d}\Theta}{\int p(D \mid \Theta)  p(\Theta) \text{d}\Theta}\\
&= \frac{ \iint p(y_*\mid x_*, \theta^{(n)}, D)\, p(D \mid \theta^{(n)})  p(\theta^{(n)},n) \text{d}\theta^{(n)}\text{dn}}{\iint p(D \mid \theta^{(n)})  p(\theta^{(n)}, n) \text{d}\theta^{(n)}\text{dn}}\\
&= \frac{ \sum_{n=1}^N p(n) \int p(y_*\mid x_*, \theta^{(n)}, D)\, p(D \mid \theta^{(n)})  p(\theta^{(n)}) \text{d}\theta^{(n)}}{\sum_{n=1}^N p(n) \int p(D \mid \theta^{(n)})  p(\theta^{(n)}) \text{d}\theta^{(n)}}, 
\end{split}
\end{equation}
where $\theta^{(n)}$ is the description of a GMM with exactly $n$ components. For our experiments we choose uniform priors over $n$, which cancel (note that this is not an inherent limitation of our method).

A na\"ive approach performs the integrals over $\theta^{(n)}$ separately before computing the sums over $n$.
The ability of MASKERADE to share information across models enables the selection of acquisitions which are maximally informative about the \textit{sum of integrals rather than each individual integral separately}, improving the convergence rate.
Additionally, accounting for the covariance between integrals over $\theta^{(n)}$ improves the quality of the posterior over the result of the summations.

\subsection{DERIVATION OF INFORMATION THEORETIC SAMPLE ACQUISITION}\label{app:acquisition}
The acquisition function is the expected information gained about the integral by making a set of likelihood observations
\begin{equation*}
	\alpha(\theta^{(n)}_{*}) = \int \bigl( H[Z \mid z, \Theta] - H[Z \mid z_{*}, \theta^{(n)}_{*}, z, \Theta] \bigr) p(z_{*} \mid \theta^{(n)}_{*}, z, \Theta) \mathrm{d} z_{*}
\end{equation*}
Due to the symmetry of the mutual information, we can swap the role of $Z$ and $z_{*}$ [52]
\begin{align*}
	\alpha(\theta^{(n)}_{*}) &= H[z_{*} \mid z, \Theta] - \int H[z_{*} \mid \theta^{(n)}_{*}, Z, z, \Theta] p(Z \mid z, \Theta) \mathrm{d} Z \\
	&= H[z_{*} \mid z, \Theta] - H[z_{*} \mid \theta^{(n)}_{*}, Z, z, \Theta]
\end{align*}
where $H$ is differential entropy (always of Gaussians).
The second line follows because the entropy of the Gaussian $p(z_{*} \mid \theta^{(n)}_{*}, Z, z, \Theta)$ does not depend on the value of $Z$.

\subsection{Monte Carlo approximation to the kernel integrals}
The kernel integrals are approximated using Quasi Monte Carlo sampling under the prior.
Table \ref{tab:mc} shows the effect of varying the number of QMC samples.

\begin{table}
\caption{The effect of number of Monte Carlo samples used to estimate the kernel integrals (i.e. the those described in Section \ref{WSABI}).
For a MASKERADE model that marginalises over up to 5 mixture components we randomly sample sets of 100, 500 and 1000 hyperparameters from the prior, and compute their corresponding likelihoods on the UCI Airfoil Self-Noise dataset.
We then infer the model evidence with WSABI using the MASKERADE hyper-kernel.
For a given number of MC samples, we repeat this five times and report the mean and SEM for the posterior mean of the model evidence.
We observe that the estimate for the model evidence is not highly sensitive to the number of monte carlo samples used to compute the kernel integrals.}
\vskip 0.15in
\begin{center}
\begin{small}
\begin{sc}
\begin{tabular}{lccc}
\toprule
& \multicolumn{3}{c}{No. of MC samples} \\
\cmidrule(lr){2-4}
No. Likelihood Observations & 100 & 1000 & 10000 \\
\midrule
100 & $0.02546 \pm 0.00149$ & $0.02542 \pm 0.00020$ & $0.02526 \pm 0.00012$ \\
500 & $0.00369 \pm 0.00035$ & $0.00379 \pm 0.00013$ & $0.00358 \pm 0.00003$ \\
1000 & $0.00333 \pm 0.00095$ & $0.00292 \pm 0.00031$ & $0.00292 \pm 0.00006$ \\
\bottomrule
\end{tabular}
\end{sc}
\end{small}
\end{center}
\label{tab:mc}
\vskip -0.1in
\end{table}

\section{FULL ALGORITHM DESCRIPTION}
For completeness we once more give the full generative model:
\begin{align}\label{model_specification}
\begin{split}
n &\sim\, \text{Uniform}(\{1,2,..N\}), \\
w &\sim\, \text{Dirichlet}(\alpha), \\
m_{1..n} &\sim\, \mathcal{N}(\mu, \Sigma),\\
\sigma_{1..n} &\sim\, \text{Log-Normal}(\nu, \tau),\\
S(\omega) &=\, \sum_{j=1}^n \frac{w_j}{2} \bigl( \mathcal{N}(\omega;m_j,\sigma_j) + \mathcal{N}(\omega;-m_j,\sigma_j) \bigr),\\
k(x, x') &=\, \int \text{e}^{2\pi i |x - x'|\omega}S(\omega) \text{d}\omega,\\
f &\sim\,  \mathcal{GP}(0, k).
\end{split}
\end{align}

\begin{figure}
    \centering
    \includegraphics[width=0.8\textwidth]{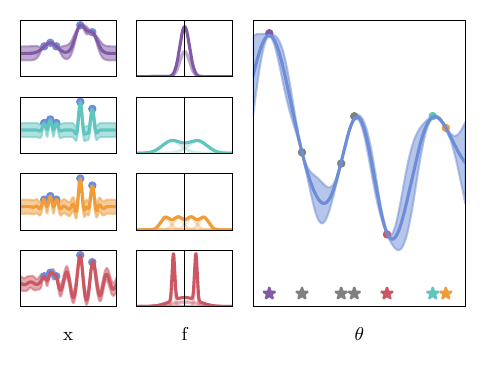}
    \caption{A schematic representation of the inference procedure under our model.  The left column shows the GP posterior for different Spectral Mixture Kernels on the same dataset. The centre column shows the SM kernels in their spectral domain. The right column illustrates a GP posterior on a space indexed by the spectral densities of the SM kernels. BQ can then be used to marginalise over SM kernels which may have different numbers of mixture components.}
    \label{fig:schematic}
\end{figure}

Below, we provide a summary of the MASKERADE framework in Algorithm 1. A schematic outlining our method is made available in Figure \ref{fig:schematic}.

\begin{algorithm}
\caption{Pseudocode for both the learning and prediction phases of our algorithm.}
\begin{algorithmic}\label{alg}
\State obtain initial samples $\Theta, z$ \Comment{$z$ are likelihood evaluations at $\Theta$.}
\State $\lambda, l \gets \mathrm{argmax}_{\lambda, l} p(z \mid \Theta, \lambda, l)$ \Comment{Optimise BQ Surrogate}
\While{$i \geq 0$}
	\State $\{\theta^{(n)}_{s}\}_{1:b} \gets \mathrm{argmax}_{\{\theta^{(n)}_{s}\}_{1:b}} \alpha(\{\theta^{(n)}_{s}\}_{1:b})$ \Comment{Optimise acquisition function}
	\State append $\{\theta^{(n)}_{s}\}_{1:b}$ to $\Theta$.
	\State append $\mathrm{likelihood}(\{\theta^{(n)}_{s}\}_{1:b})$ to $z$.
	\State $\lambda, l \gets \mathrm{argmax}_{\lambda, l} p(z \mid \Theta, \lambda, l)$ \Comment{Optimise BQ surrogate}
	\State $i \gets i - 1$
\EndWhile

\State $Q \gets \mathrm{compute\_bq\_weights}(\Theta, z, \lambda, l, \phi)$ \Comment{$Q$ defined as in Appendix on WSABI above. $\phi$ are all prior parameters.}

\For{$\theta^{(n)}_s$ in $\Theta$}
\State $\mu_{*}, \Sigma_{*} \gets \mathrm{predict}(x_{*}, D, \theta^{(n)}_s)$
\EndFor

\Return $\mathrm{predictive\_posterior}(Q, \{\mu_{*}\}, \{\Sigma_{*}\}, z)$ \Comment{See Appendix on WSABI for details.}
\end{algorithmic}
\end{algorithm}

\section{QUALITATIVE ANALYSIS}
\begin{figure}[h]
\centering
\begin{subfigure}[b]{0.3\linewidth}
 \centering
 \includegraphics[width=\linewidth]{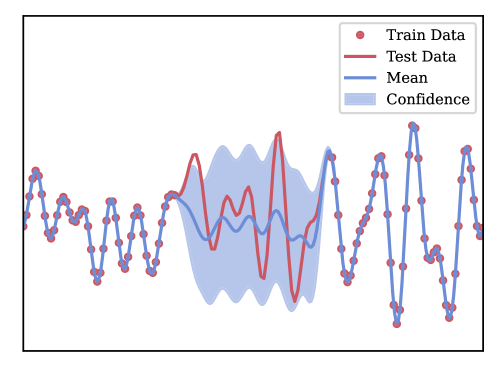}
 \label{fig:sm_posterior}
\end{subfigure}
\hfill
\begin{subfigure}[b]{0.3\linewidth}
 \centering
 \includegraphics[width=\linewidth]{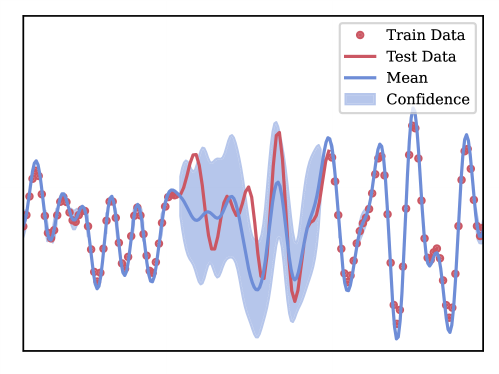}
 \label{fig:bq3_posterior}
\end{subfigure}
\hfill
\begin{subfigure}[b]{0.3\linewidth}
 \centering
 \includegraphics[width=\linewidth]{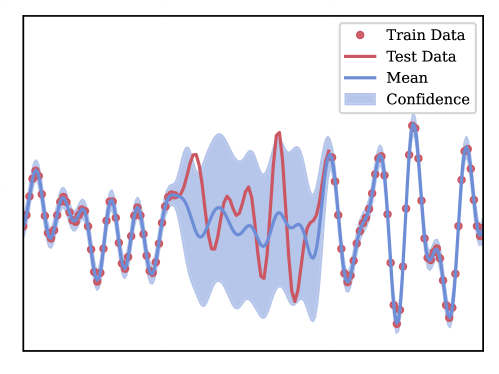}
 \label{fig:bq8_posterior}
\end{subfigure}

\begin{subfigure}[b]{0.3\linewidth}
 \centering
 \includegraphics[width=\linewidth]{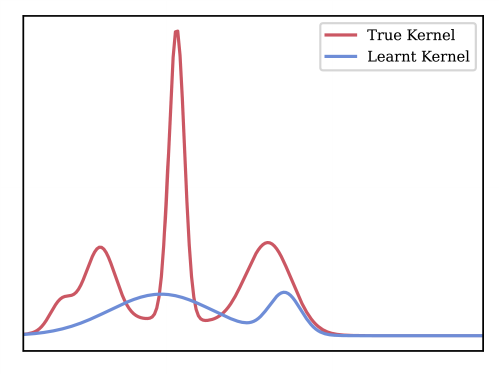}
 \caption{SM Kernel}
 \label{fig:sm_kernel}
\end{subfigure}
\hfill
\begin{subfigure}[b]{0.3\linewidth}
 \centering
 \includegraphics[width=\linewidth]{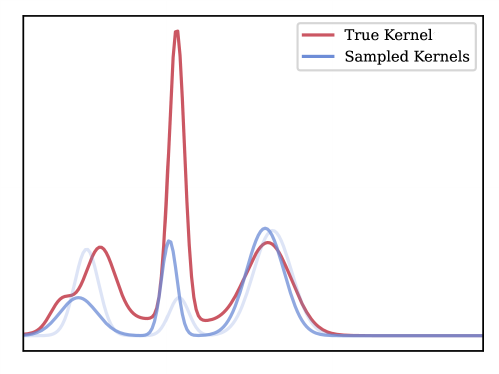}
 \caption{MASKERADE 1--3}
 \label{fig:bq3_kernel}
\end{subfigure}
\hfill
\begin{subfigure}[b]{0.3\linewidth}
 \centering
 \includegraphics[width=\linewidth]{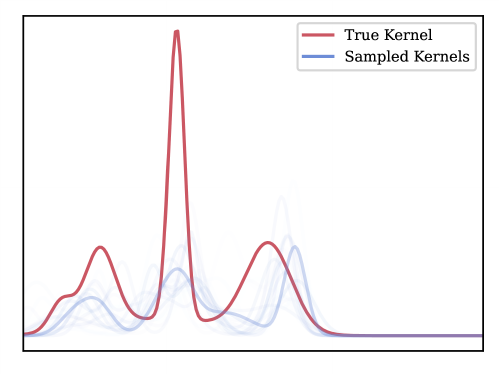}
 \caption{MASKERADE 1--8}
 \label{fig:bq8_kernel}
\end{subfigure}
\caption{A comparison of a 5 component SM kernel with optimised hyperparameters, and two variants of MASKERADE -- one that places a prior over up to 3 mixture components, and the other up to 8 mixture components -- on a toy dataset drawn from a 5 component SM kernel.
Each column plots attributes of the labelled model.
The first row shows the posterior conditioned on the training data (for the MASKERADE models we show the moment-matched posterior).
The second row shows the spectra (for positive frequencies) of the data generating kernel, and (for the 5 component SM kernel model) the optimised or (for the MASKERADE models) the sampled kernels.
For the MASKERADE model, the opacity of a sampled kernel is proportional to the quadrature weight for all GP products of which that kernel is a part.
(Recall that the posterior is a weighted sum of products of GP posteriors.)
MASKERADE 1--3 is able to select samples near the data generating kernel, and therefore produce a posterior that generalises better than the other two models.
Despite having the same number of mixture components as the data generating kernel, the optimised SM kernel sets the weights of 3 components to be very small.
MASKERADE 1--8 struggles to explore its larger hyperparameter space with the same budget (500 likelihood evaluations) as MASKERADE 1--3.
This can be seen by the fact that it spreads its posterior mass more evenly over a larger number of samples.}
\label{fig:post_kern}
\end{figure}

We qualitatively inspect the behaviour of MASKERADE in Figure \ref{fig:post_kern} by examining the spectra of kernels assigned the highest weights in the posterior, and the effect of varying the number of mixture components that are marginalised over.

\begin{figure}
     \centering
     \begin{subfigure}[b]{0.32\textwidth}
         \centering
         \includegraphics[width=\textwidth]{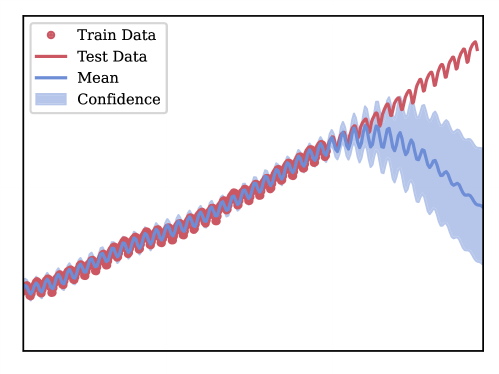}
         \caption{MASKERADE}
         \label{fig:mlmas}
     \end{subfigure}
     \hfill
     \begin{subfigure}[b]{0.32\textwidth}
         \centering
         \includegraphics[width=\textwidth]{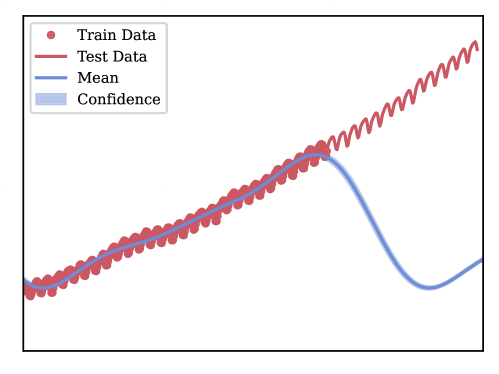}
         \caption{FKL}
         \label{fig:mlfkl}
     \end{subfigure}
     \hfill
     \begin{subfigure}[b]{0.32\textwidth}
         \centering
         \includegraphics[width=\textwidth]{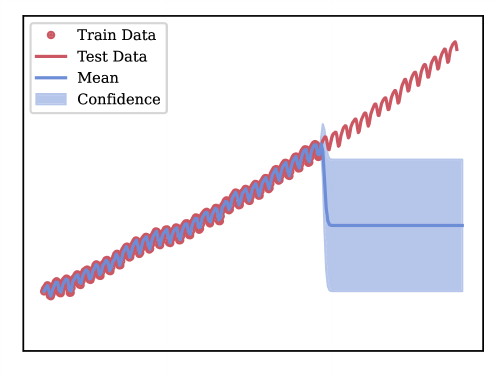}
         \caption{Spectral Mixture Kernel}
         \label{fig:mlsm}
     \end{subfigure}
        \caption{Posteriors for several methods on the Mauna Loa dataset (For MASKERADE and FKL we show moment matched posteriors). All methods struggle to model the linear trend, but MASKERADE is best able to extrapolate the periodic structure.}
        \label{fig:mauna_loa}
\end{figure}

Next, we visually verify that our method improves upon baselines by plotting fits on the Mauna Loa dataset.
These are shown in Figure \ref{fig:mauna_loa}.
MASKERADE is parameterised to marginalise over 5 mixture components, with priors set as in Section 3.1.1.
FKL uses the defaults in the \texttt{spectralgp}\footnote{https://github.com/wjmaddox/spectralgp} Python package, with the maximum frequency set to the Nyquist frequency of the Mauna Loa dataset.
The Spectral Mixture kernel uses 5 mixtures, initialised using the empirical spectrum of the dataset.

